\documentclass[pdflatex,sn-mathphys-num]{sn-jnl}

\usepackage{graphicx}%
\usepackage{multirow}%
\usepackage{amsmath,amssymb,amsfonts}%
\usepackage{amsthm}%
\usepackage{mathrsfs}%
\usepackage[title]{appendix}%
\usepackage{xcolor}%
\usepackage{textcomp}%
\usepackage{manyfoot}%
\usepackage{booktabs}%
\usepackage{algorithm}%
\usepackage{algorithmicx}%
\usepackage{algpseudocode}%
\usepackage{listings}%

\usepackage{tikz}
\usetikzlibrary{svg.path}
\definecolor{orcidlogocol}{HTML}{A6CE39}
\tikzset{
  orcidlogo/.pic={
    \fill[orcidlogocol] svg{M256,128c0,70.7-57.3,128-128,128C57.3,256,0,198.7,0,128C0,57.3,57.3,0,128,0C198.7,0,256,57.3,256,128z};
    \fill[white] svg{M86.3,186.2H70.9V79.1h15.4v48.4V186.2z}
                 svg{M108.9,79.1h41.6c39.6,0,57,28.3,57,53.6c0,27.5-21.5,53.6-56.8,53.6h-41.8V79.1z M124.3,172.4h24.5c34.9,0,42.9-26.5,42.9-39.7c0-21.5-13.7-39.7-43.7-39.7h-23.7V172.4z}
                 svg{M88.7,56.8c0,5.5-4.5,10-10,10c-5.5,0-10-4.5-10-10c0-5.5,4.5-10,10-10C84.2,46.8,88.7,51.3,88.7,56.8z};
  }
}
\newcommand{\sorcidlink}[1]{\href{https://orcid.org/#1}{\raisebox{-1pt}{\tikz[baseline=-0.6ex]{\pic[scale=0.024]{orcidlogo};}}}}

\theoremstyle{thmstyleone}%

\theoremstyle{thmstyletwo}%

\theoremstyle{thmstylethree}%

\begin{document}

\title[SAPE: Sandwich Adapters for LLM Fine-Tuning]{SAPE: Sandwich Adapters for Parameter Efficiency in Large Language Model Fine-Tuning}

\author[1]{\fnm{Mohammad Aref} \sur{Jafari-Raddani}\,\sorcidlink{0009-0001-6052-0062}}\email{raddaniaref@gmail.com}

\author*[1]{\fnm{Morteza} \sur{Mohajjel Kafshdooz}\,\sorcidlink{0000-0002-0303-1238}}\email{mohajjel@qut.ac.ir}

\affil*[1]{\orgdiv{Department of Electrical and Computer Engineering}, \orgname{Qom University of Technology}, \orgaddress{\city{Qom}, \country{Iran}}}

\abstract{
While Parameter-Efficient Fine-Tuning (PEFT) has substantially reduced the hardware cost of adapting Large Language Models (LLMs) by decreasing the number of trainable parameters, recent studies have sought to further improve PEFT through parameter sharing. However, these approaches either employ uniform parameter sharing across layers, which can delay convergence, or rely on dynamic masking strategies, which add computational overhead. The potential of sharing patterns inspired by the inherent hierarchical structure of Transformer architectures remains unexplored in PEFT. To address this gap, we introduce SAPE (Sandwich Adapters for Parameter Efficiency), a PEFT framework based on a sandwich-style hard weight-sharing topology. SAPE routes intermediate Transformer layers through balanced shared group adapters while strictly isolating the input embedding and final projection boundary transformations to prevent gradient interference. This design significantly reduces memory consumption while eliminating the computational overhead associated with dynamic parameter-sharing methods. Extensive evaluations across encoder-only and causal decoder architectures demonstrate that SAPE achieves state-of-the-art performance in low-parameter regimes. On natural language understanding, SAPE outperforms proPETL on RoBERTa-large while utilizing only 10\% of the baseline's parameter budget. On natural language generation and world knowledge reasoning with LLaMA-3.2 (3B) under a strict $\sim$0.6M parameter constraint, SAPE outperforms AdaLoRA, yielding absolute improvements of +4.85\% on GSM8K and +3.11\% on CommonsenseQA. Furthermore, through comprehensive topological ablations, we formalize an inherent capacity trade-off: while hard parameter sharing strongly regularizes semantic generalization, it slightly smooths the sharp layer-wise transformations required for rigid multi-step arithmetic reasoning. Additionally, ablation studies confirm the robustness of SAPE's architectural design choices.
}

\keywords{Large Language Models, Parameter-Efficient Fine-Tuning, Adapter Tuning, Transformer Architectures, Weight Sharing, sandwich Topology}

\maketitle

\section{Introduction}\label{sec1}

The advent of Transformer-based pre-trained models has fundamentally revolutionized Natural Language Processing (NLP). Fine-tuning these architectures has yielded state-of-the-art performance across a wide range of downstream tasks  \citep{han2024parameter, raiaan2024review}, a breakthrough largely attributed to their immense capacity—enabled by billions of parameters—and pre-training on vast datasets. However, the sheer scale of these models renders full fine-tuning prohibitively expensive in terms of both memory and computational resources.

To mitigate the computational and memory costs imposed by large-scale Transformer models, Parameter-Efficient Fine-Tuning (PEFT) techniques have been proposed as an effective alternative to full fine-tuning. Adapter tuning, as one of the earlier PEFT methods, was introduced by \citet{houlsby2019parameter} and \citet{pfeiffer2021adapterfusion}, which involves training a small, non-linear bottleneck inserted sequentially into feed-forward sublayers within Transformer layers. Additionally, Low-Rank Adaptation (LoRA) \citep{hu2021lora} was proposed as another PEFT approach that reparameterizes weight updates as low-rank  matrices. Although these methods are highly successful in reducing memory costs, there is a noticeable performance gap between them and the full fine-tuning strategy \citep{biderman2024lora}. To address the rank collapse problem in LoRA, AdaLoRA \citep{zhang2023adalora} was proposed to dynamically allocate the parameter budget across different weight matrices based on their relative importance, instead of applying a fixed rank uniformly. However, this dynamic rank allocation relies on continuous importance scoring and iterative masking during training, which introduces a computational overhead.

A parallel line of research has demonstrated that cross-layer parameter sharing serves as a regularizer and simultaneously reduces the memory footprint of large-scale Transformer models. For example, ALBERT \citep{lan2019albert} demonstrated the viability of sharing entire Transformer blocks to improve parameter efficiency. Inspired by this idea, the Subformer architecture \citep{reid2021subformer} refined this approach for generative modeling by introducing a ``sandwich-style'' topology, where intermediate layers share parameters while crucial boundary layers remain independent to preserve representational power. Furthermore, \citet{pires2023one} demonstrated that sharing a single, wide Feed-Forward Network (FFN) across encoder layers yields superior accuracy and latency compared to standard, isolated narrow networks.

Inspired by these parameter-sharing techniques, recent studies have begun integrating cross-layer parameter sharing into PEFT methods. For instance, ShareLoRA \citep{song2024sharelora} and Tied-LoRA \citep{renduchintala2024tied} demonstrated that sharing low-rank projection matrices uniformly across all layers significantly reduces the parameter footprint compared to standard LoRA. However, this homogeneous sharing strategy leads to delayed convergence. Nevertheless, while the LoRA paradigm has seen extensive exploration in parameter sharing, a significant research gap remains: cross-layer sharing within the Adapter paradigm is remarkably underexplored. The few existing attempts, such as proPETL \citep{zeng2023one}, share a single prototype network across all Transformer blocks, relying on learnable binary masks to dynamically prune connections. Nonetheless, this masking step introduces a slight overhead during training. Concurrently, within Automatic Speech Recognition (ASR), methods like Hierarchical Recurrent Adapters (HRA) \citep{munkhdalai2024hierarchical} and Shared-Adapters \citep{rolland2024shared} have successfully implemented cross-layer adapter sharing in speech models. However, these ASR approaches also utilize homogeneous sharing. Despite the efficacy of the sandwich-style topology in layer sharing like Subformer \citep{reid2021subformer}, its potential application within Parameter-Efficient Fine-Tuning (PEFT) remains unexplored. Specifically, investigating a sandwich sharing configuration for the additive paradigm is an unaddressed opportunity.

\begin{figure}[h]
\centering
\includegraphics[width=0.9\textwidth]{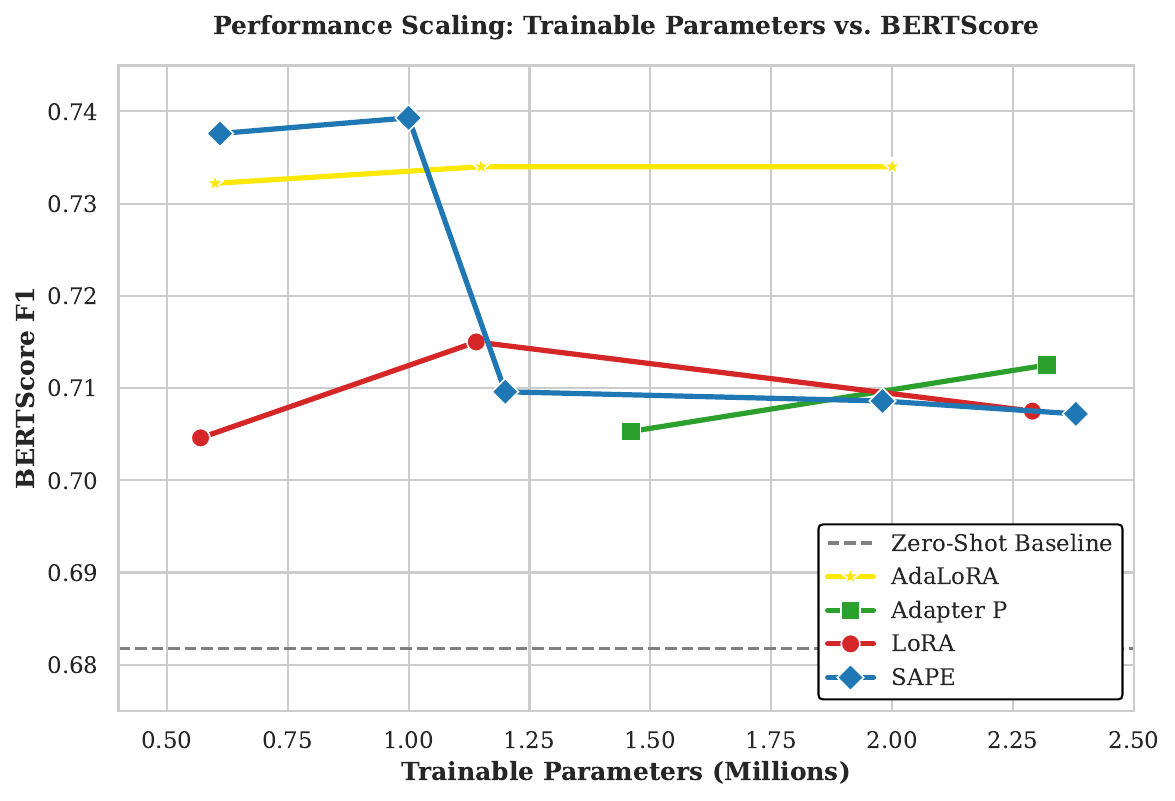}
\caption{Generative Performance vs. Trainable Parameters on ConvAI2. SAPE maintains high semantic fidelity (BERTScore) even at extreme sub-million parameter constraints.}\label{fig:convai2_efficiency}
\end{figure}

To bridge this gap, we introduce \textbf{SAPE} (\textbf{S}andwich \textbf{A}dapters for \textbf{P}arameter \textbf{E}fficiency), an additive PEFT method that employs a static, weight-sharing topology aligned with the hierarchical representation depth of Transformers. Unlike prior parameter-shared PEFT approaches that either enforce uniform layer-wise tying or rely on dynamic layer-wise masking with additional computational cost, SAPE leverages parameter sharing through a deterministic sandwich structure. Specifically, the proposed framework incorporates a dual-mechanism design: 
(i)~It isolates the critical boundary layers ($L_0$ and $L_{N-1}$) with dedicated, unshared adapters, preserving independent feature adaptation at raw embedding inputs and task-specific projections to prevent boundary gradient interference; and 
(ii)~It partitions the intermediate layers into $G$ balanced groups where adapters within each group share hard parameter weights. This deterministic topology drastically compresses trainable parameter complexity while functioning as an effective structural regularizer—completely eliminating the auxiliary memory and latency overhead associated with dynamic rank allocation or mask generation schedules.

The primary contributions of this paper are summarized as follows:

\begin{itemize}
    \item \textbf{Novel Structural PEFT Framework:} We propose SAPE, an additive PEFT framework that leverages a boundary-isolated sandwich topology for hard adapter weight sharing. SAPE eliminates parameter redundancy across intermediate layers while preventing boundary gradient interference, achieving high parameter compression without relying on computationally expensive dynamic methods.

    \item \textbf{State-of-the-Art Low-Parameter Efficiency:} We evaluate SAPE across both decoder-only and encoder-only Transformers on diverse downstream tasks, achieving state-of-the-art performance in low-parameter regimes. Notably, when fine-tuning LLaMA-3.2 (3B) with a strict budget of $\sim$0.6M parameters, SAPE outperforms AdaLoRA across major reasoning and generative metrics, yielding substantial absolute improvements of +4.85\% on GSM8K and +3.11\% on CommonsenseQA. Furthermore, SAPE outperforms proPETL Adapter on the GLUE benchmark using RoBERTa-large, despite utilizing only 10\% of the baseline's parameter budget.

    \item \textbf{Comprehensive Architectural \& Empirical Insights:} We present a thorough series of topological and mechanistic ablations, establishing four core insights into shared adaptation. First, we show that hard parameter sharing strongly regularizes semantic generalization while marginally smoothing rigid step-by-step arithmetic reasoning. Second, topological ablations confirm that boundary layer isolation ($L_0, L_{N-1}$) is structurally necessary to prevent representational collapse on complex entailment tasks. Third, we show that performance does not scale monotonically with group granularity ($G$), identifying $G=2$ as the optimal configuration for DeBERTa-v3-base, retaining over $99.4\%$ of unshared baseline performance at one-third of the parameter cost. Finally, we demonstrate that while framework performance is robust to spatial density allocation ($\text{DenseEarly}$ vs. $\text{DenseLate}$), it benefits significantly from layer-wise independent dropout masking dynamics.
\end{itemize}

\section{Related Work}\label{sec2}

\subsection{Parameter-Efficient Fine-Tuning (PEFT)}
To mitigate the high computational cost of full fine-tuning of LLMs, parameter-efficient fine-tuning methods have been proposed to reduce the number of trainable parameters while maintaining full fine-tuning performance. \citet{houlsby2019parameter} pioneered the insertion of an adapter module with a non-linear bottleneck after both the attention and feed-forward sublayers. By only training this modular adapter, not only is the number of trainable parameters reduced significantly, but the modular design also reduces the memory required to store task-specific parameters. Subsequently, \citet{pfeiffer2021adapterfusion} demonstrated that isolating a single adapter after the feed-forward sublayer is more efficient. Furthermore, \citet{he2022sparseadapter} introduced \textit{SparseAdapter} to further reduce parameter redundancy within adapter modules. They prune adapter weight matrices at initialization utilizing different pruning strategies, such as random pruning and SNIP. They demonstrate that adapters remain effective even at high sparsity levels (up to 80\%). Building on this insight, their proposed \textit{Large-Sparse} setting scales up the bottleneck dimension alongside a high sparsity ratio, boosting representation capacity while strictly maintaining a fixed parameter budget. However, while SparseAdapter improves parameter efficiency locally via unstructured intra-module pruning, it does not share parameters across Transformer layers, leaving macro-level layer redundancy unaddressed. Besides bottleneck adapters, soft prompt methods represent another additive PEFT paradigm. Pioneered by Prefix-Tuning~\citep{li2021prefix} and Prompt Tuning~\citep{lester2021power}, these methods prepend a sequence of learnable continuous vectors to input embeddings or intermediate activations, achieving high parameter efficiency while leaving all pre-trained backbone weights completely frozen.

In contrast to additive modules, selective PEFT methods modify only a small, specific subset of the model's pre-trained parameters. For instance, BitFit~\cite{zaken2022bitfit} fine-tunes solely the internal bias vectors of the network while keeping all weight matrices frozen. Despite its high parameter efficiency, it faces key limitations, such as a strict architectural dependency, and its low parameter capacity bottlenecks performance when adapting to complex domain shifts.

Another paradigm of PEFT is decomposition-based methods, which reduce the number of trainable parameters by learning low-rank weight updates that are added to the frozen pre-trained weights of the model. In this case, LoRA~\citep{hu2021lora} introduces low-rank matrices $A$ and $B$ representing the weight update $\Delta W = AB$ of the query and value projection matrices. LoRA enables augmenting the original weight matrix with a trainable low-rank update and incurs no additional inference overhead. Despite these advantages, standard LoRA allocates a uniform rank $r$ to all reparameterized matrices, ignoring the varying importance of individual weight projections. To improve rank allocation efficiency, AdaLoRA~\citep{zhang2023adalora} utilizes an SVD-style formulation $\Delta W = P \Lambda Q$, where singular values in $\Lambda$ are dynamically pruned during training based on sensitivity-based importance scoring. Building upon this paradigm, subsequent works, including DyLoRA~\citep{valipour2023dylora} and DoRA~\citep{liu2024dora}, explore dynamic rank selection and weight-decomposed adaptation, respectively. (We refer readers to \citet{wang2025parameter} for a comprehensive survey on all these PEFT methods).

Although AdaLoRA efficiently determines rank allocation, it introduces minor computational overhead due to iterative importance scoring and dynamic pruning schedules. Nevertheless, despite the parameter efficiency of the LoRA paradigm, \citet{ko2025lora} recently demonstrated that LoRA can sometimes be even slower than full fine-tuning. This occurs because GPUs process a single kernel at a time; inserting parallel low-rank adapter branches ($A$ and $B$) forces the execution of sequential CUDA kernels, creating a computational bottleneck.

\subsection{Cross-Layer Parameter Sharing}
To reduce the overall size of large language models and improve pretraining and memory usage efficiency, parameter sharing across layers has been explored. ALBERT~\citep{lan2019albert} utilizes cross-layer parameter sharing in BERT to eliminate the increase in parameters caused by growing the depth of the network. By gradually increasing the hidden size, they illustrate that it is possible to surpass BERT even with a lower number of parameters. However, the trade-off is between the overhead of training time due to larger width and better performance and memory consumption. Building upon these parameter-sharing paradigms, Subformer~\citep{reid2021subformer} suggests a sandwich-style sharing strategy, which shares middle layers $L_1$ to $L_{N-2}$ as one group with a wider dimension than the base model dimension, and keeps two separate layers for $L_0$ and $L_{N-1}$ as they interact with raw embeddings and final outputs. This structure-wise sharing not only outperforms uniform sharing across all layers, but also reaches comparable performance compared to a base model with more parameters, achieving even better performance and faster training convergence in some tasks. However, the trade-off of increasing the dimension of the shared layers remains. Despite these successful results, structure-wise sharing is still a gap in fine-tuning methods.

\subsection{Parameter Sharing within PEFT}

 In an early foundational work, \textit{Compacter}~\citep{karimi2021compacter} introduced parameter sharing in bottleneck adapters by leveraging Parameterized Hypercomplex Multiplication (PHM) layers to share global factor matrices $A_i$ across all Transformer layers, while employing layer-specific rank-one weight factorizations ($s_i t_i^\top$) for adaptation. Building on the concept of shared prototypes in PEFT, \citet{zeng2023one} introduced the proPETL framework. proPETL trains a shared prototype network such as an adapter module, LoRA parameters, or a prefix across all layers, applying layer-specific binary masks to extract individual sub-networks. Notably, their findings demonstrate that utilizing an adapter as the shared prototype outperforms LoRA or prefix sharing, although the requirement for layer-wise mask search and storage introduces non-trivial memory and computational overhead. Additionally, while factorized and masked adapter sharing have been investigated, direct, macro-level hard weight sharing of complete bottleneck adapters across structured layer groups remains remarkably underexplored.

 In parallel, ShareLoRA~\citep{song2024sharelora} explores a variety of sharing approaches in LoRA, including sharing only the $A$ matrix with a unique $B$ matrix per layer, sharing only $B$ with an independent $A$ matrix, or sharing both at the same time. Despite competitive results with shared $A$, configuration variants that share $B$ or both matrices tend to converge more slowly and underperform. Recently, MASA~\citep{dong2025masa} leveraged cross-layer parameter sharing in multi-task learning by proposing a LoRA-based framework. Instead of tying parameters across all layers uniformly like ShareLoRA~\citep{song2024sharelora}, MASA shares an ensemble of $A$ matrices across adjacent layer groups while retaining layer-specific $B$ matrices. Furthermore, rather than relying on dynamic routing mechanisms such as AdapterFusion~\citep{pfeiffer2021adapterfusion}, MASA aggregates the outputs of these $A$ matrices via simple summation. These asymmetric, heterogeneous sharing strategies yield superior multi-task adaptation capabilities while maintaining parameter efficiency.
 
However, all of these parameter sharing strategies across both additive adapters and LoRA reparameterizations enforce parameter tying uniformly across all depth levels, thereby overlooking depth-wise structural sharing patterns. SAPE tackles this unaddressed gap by defining a heterogeneous topology based on sandwich sharing for adapters.

\section{Methodology}\label{sec:methodology}

As an adapter-based PEFT framework, SAPE employs both shared and isolated adapters across distinct Transformer layers. This section begins by defining the architecture of the base adapter. Next, driven by the hierarchical architecture of Language Models (LMs), the proposed sandwich sharing topology is detailed. Finally, the optimization routing for this paradigm is formulated, in addition to formalizing the total trainable parameter complexity.

\begin{figure*}[h]
\centering
\includegraphics[width=\textwidth]{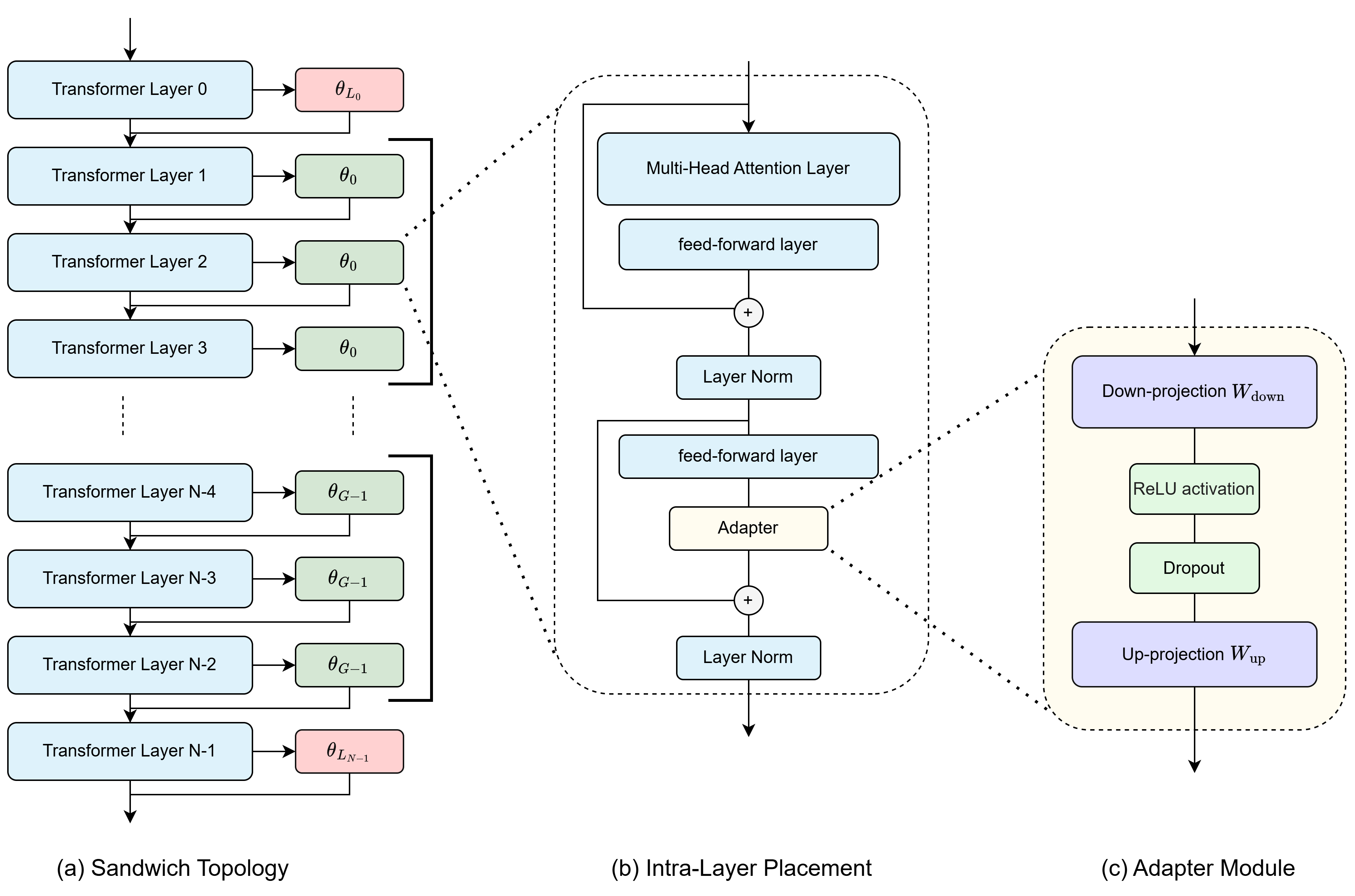}
\caption{\textbf{Architectural overview of the SAPE framework.} The diagram illustrates the macro-to-micro structural constraints of the proposed sandwich topology. 
\textbf{(a) Sandwich Topology:} Network-level parameter sharing across $N$ Transformer layers, depicting isolated boundary adapters ($\mathcal{A}(0) = \theta_{L_0}, \mathcal{A}(N-1) = \theta_{L_{N-1}}$) alongside shared intermediate groups ($\mathcal{I}_g \implies \mathcal{A}(l) = \theta_g$). 
\textbf{(b) Intra-Layer Placement:} Detailing the injection of the bottleneck module $\mathcal{F}_\theta$ immediately following the final feed-forward sub-layer, receiving the hidden state $\mathbf{h}_l$. 
\textbf{(c) Adapter Module:} Internal tensor operations of the bottleneck adapter, highlighting the down-projection ($\mathbf{W}_{\text{down}} \in \mathbb{R}^{d \times m}$), non-linear ReLU activation $\sigma(\cdot)$, dropout regularization, and up-projection ($\mathbf{W}_{\text{up}} \in \mathbb{R}^{m \times d}$).}\label{fig:sape_architecture}
\end{figure*}

\subsection{Adapter Architecture and Placement}

SAPE utilizes a bottleneck architecture with non-linear activation, similar to the adapter modules which were introduced by \citet{houlsby2019parameter} and the optimized single-adapter configuration by \cite{pfeiffer2021adapterfusion}. Let $\mathbf{h}_l \in \mathbb{R}^d$ denote the hidden state output of the final feed-forward sub-layer at the $l$-th Transformer layer. The adapter transformation is parameterized by $\theta = \{\mathbf{W}_{down}, \mathbf{W}_{up}, \mathbf{b}_{down}, \mathbf{b}_{up}\}$ and is formalized as the mapping $\mathcal{F}_{\theta}: \mathbb{R}^d \rightarrow \mathbb{R}^d$:

\begin{equation}
\mathcal{F}_{\theta}(\mathbf{h}_l) = \mathbf{h}_l + \Phi \Big( \sigma(\mathbf{h}_l \mathbf{W}_{down} + \mathbf{b}_{down}) \Big) \mathbf{W}_{up} + \mathbf{b}_{up}
\end{equation}

where $\mathbf{W}_{down} \in \mathbb{R}^{d \times m}$ projects the $d$-dimensional input to a smaller bottleneck dimension $m$ ($m \ll d$), and $\mathbf{W}_{up} \in \mathbb{R}^{m \times d}$ projects it back to the original dimension. The function $\sigma(\cdot)$ represents the ReLU activation, and $\Phi(\cdot)$ denotes a dropout operator with a rate of $10\%$ applied during training to mitigate overfitting. In the SAPE framework, $\mathcal{F}_{\theta}$ is applied once per Transformer layer after the last feed-forward layer.

We employ a hybrid adapter injection strategy that structurally separates boundary transformations from intermediate semantic routing. Inspired by the sandwich-style weight sharing paradigm introduced by \citet{reid2021subformer} for generative models, we partition the entire set of Transformer layers $\mathcal{L}$ into a set of boundary layers $\mathcal{B}$ and a set of intermediate layers $\mathcal{I}$:

\begin{equation}
\mathcal{B} = \{0, N-1\}, \quad \mathcal{I} = \mathcal{L} \setminus \mathcal{B} = \{1, 2, \dots, N-2\}
\end{equation}

To implement the sandwich topology, we define an adapter allocation function $\mathcal{A}: \mathcal{L} \rightarrow \boldsymbol{\Theta}$, mapping each layer $l$ to a specific parameter set $\theta \in \boldsymbol{\Theta}$. For the critical boundary layers, which capture low-level syntax and high-level task-specific semantics respectively, as corroborated by extensive probing studies~\citep{jawahar2019does, tenney2019bert, vulic2020probing}, we allocate strictly isolated parameter sets:

\begin{equation}
\mathcal{A}(0) = \theta_{L_0}, \quad \mathcal{A}(N-1) = \theta_{L_{N-1}}
\end{equation}

For the intermediate layers $\mathcal{I}$, we partition the set into $G$ disjoint subsets (groups), such that $\mathcal{I} = \bigcup_{g=0}^{G-1} \mathcal{I}_g$ and $\mathcal{I}_i \cap \mathcal{I}_j = \emptyset$ for $i \neq j$. All layers within a subset $\mathcal{I}_g$ share a single adapter parameter set $\theta_{g}$:

\begin{equation}
\forall l \in \mathcal{I}_g \implies \mathcal{A}(l) = \theta_g
\end{equation}

After partitioning the intermediate layers into $G$ groups, the objective is to distribute the $N-2$ layers as uniformly as possible among them. Let

\begin{equation}
s = \left\lfloor \frac{N - 2}{G} \right\rfloor, \quad r = (N - 2) \bmod G,
\end{equation}

where $s$ denotes the minimum number of layers assigned to each group, and $r$ represents the number of remaining layers after the uniform allocation. Consequently, each group contains either $s$ or $s+1$ layers, ensuring that the size difference between any two groups is at most one layer.

The distribution of the $r$ remaining layers is controlled by the boolean hyperparameter \textit{DenseEarly}. When \textit{DenseEarly} = True, the additional layers are assigned to the first $r$ groups, resulting in denser grouping toward the earlier Transformer layers. Conversely, when \textit{DenseEarly} = False, the additional layers are assigned to the last $r$ groups, producing denser grouping toward the later Transformer layers. Formally, the cardinality of each group is defined as:

\begin{equation}
|\mathcal{I}_g| = 
\begin{cases} 
s + 1, & \text{if } \textit{DenseEarly} = \text{True } \land \ g < r, \\
s, & \text{if } \textit{DenseEarly} = \text{True } \land \ g \ge r, \\
s + 1, & \text{if } \textit{DenseEarly} = \text{False } \land \ g \ge G - r, \\
s, & \text{if } \textit{DenseEarly} = \text{False } \land \ g < G - r.
\end{cases}
\end{equation}

This allocation guarantees that every intermediate layer is assigned to exactly one group while maintaining balanced group sizes. As a result, the sandwich topology preserves parameter sharing across the intermediate layers while allowing explicit control over whether the larger groups are concentrated near the input or near the output of Transformer.

\subsection{Optimization and Parameter Complexity}

During backpropagation, hard parameter sharing modifies the optimization dynamics by coupling the gradient updates across multiple Transformer layers. SAPE utilizes a dropout layer in all adapter modules, meaning an independent dropout mask $\mathbf{m}_l \sim \text{Bernoulli}(1-p)$ is dynamically generated during the forward pass, regardless of whether the module belongs to a shared group or an isolated boundary layer. The adapter transformation at any specific layer is formally conditioned on its local mask:

\begin{equation}
\mathcal{F}_{\theta_g}(\mathbf{h}_l; \mathbf{m}_l) = \mathbf{h}_l + \Big( \mathbf{m}_l \odot \sigma(\mathbf{h}_l \mathbf{W}_{down} + \mathbf{b}_{down}) \Big) \mathbf{W}_{up} + \mathbf{b}_{up}
\end{equation}

where $\odot$ denotes the element-wise product. Let $\mathcal{J}(\Theta)$ denote the task-specific loss function. Since the adapter parameter set $\theta_g$ is shared by all layers in the group $\mathcal{I}_g$, its gradient is obtained by accumulating the contributions from every layer in which it is applied. By the multivariate chain rule, the parameter update is given by:

\begin{equation}
\theta_g \leftarrow \theta_g - \eta \sum_{l \in \mathcal{I}_g} \left( \frac{\partial \mathcal{F}_{\theta_g}(\mathbf{h}_l; \mathbf{m}_l)}{\partial \theta_g} \right)^T \nabla_{\mathcal{F}_{\theta_g}(\mathbf{h}_l; \mathbf{m}_l)} \mathcal{J}
\end{equation}

where $\eta$ denotes the learning rate. Unlike conventional adapters, where each parameter set is optimized using the gradient of a single Transformer layer, the shared adapter receives the aggregated gradient from all layers within its assigned group. Consequently, each update is jointly influenced by the representations learned at multiple depths of the network, encouraging the shared parameters to capture features that are beneficial across the entire group rather than adapting to a single layer.

From a parameterization perspective, hard parameter sharing introduces an equality constraint among the adapters assigned to the layers in $\mathcal{I}_g$, i.e., all layers in the group are required to use the same parameter set $\theta_g$. This constraint reduces the number of trainable parameters while acting as a form of structural regularization. Because the shared adapter must accommodate the feature distributions of multiple Transformer layers simultaneously, it is encouraged to learn more generalizable transformations instead of becoming highly specialized for the representation of any individual layer. As a result, the proposed sandwich topology improves parameter efficiency while preserving the expressive capacity required for effective adaptation.

Given a bottleneck dimension $m$ and a base model dimension $d$, the total number of trainable parameters $P_{total}$ introduced by the SAPE method across the $G$ intermediate groups and $2$ boundary layers is strictly constrained to:

\begin{equation}
P_{total} = (2 + G) \times (2md + m + d)
\end{equation}

This rigorous parameterization highlights SAPE's ability to drastically compress the fine-tuning memory footprint while explicitly defining the routing and optimization pathways.

\section{Experiments}\label{sec:experiments}

The SAPE framework is implemented for fine-tuning RoBERTa-large~\citep{liu2019roberta} and DeBERTa V3-base~\citep{he2020deberta} on the General Language Understanding Evaluation (GLUE) benchmark~\citep{wang2018glue}. In addition, for natural language generation and reasoning, LLaMA-3.2 (3B)~\citep{grattafiori2024llama} is fine-tuned on a wide range of tasks utilizing SAPE and standard baselines. Finally, we conduct a comprehensive ablation study to evaluate the core design choices within SAPE, including the sandwich placement topology, group granularity ($G$), spatial density allocation($DenseEarly$), and independent dropout masking.

\subsection{Experimental Setup}
\label{subsec:setup}

\textbf{Implementation Details:} PyTorch and the Hugging Face \texttt{transformers} library are used for the implementation. We evaluate SAPE by setting \texttt{DenseEarly = False} as the default configuration across the various setups of intermediate groups ($G$) and bottleneck dimensions ($D$), as specified in each section. The experiments were conducted on an NVIDIA RTX 3090 and an NVIDIA T4 equipped with 24 GB and 15 GB of VRAM, respectively. Notably, the complete training hyperparameter configurations for all subsequent sections are detailed in Appendix~\ref{sec:appendix_hyperparameters}.

\paragraph{Datasets} 
We evaluate SAPE across the following benchmarks:
\begin{itemize}
\item \textbf{GLUE Benchmark~\citep{wang2018glue}:} Selected to assess standard Natural Language Understanding (NLU) on encoder-only architectures.
\item \textbf{ConvAI2~\citep{dinan2019second}:} A multi-turn, non-goal-oriented dialogue dataset explicitly designed to evaluate conversational consistency and persona maintenance in open-domain chatbots.
\item \textbf{CommonsenseQA~\citep{talmor2019commonsenseqa}:} A multiple-choice question answering dataset designed to evaluate question answering with prior world knowledge across complex semantic relationships.
\item \textbf{GSM8K~\citep{cobbe2021training}:} A dataset of 8.5K high-quality, linguistically diverse grade school math word problems designed to evaluate multi-step mathematical reasoning.
\end{itemize}

\paragraph{Baselines} 
SAPE is compared with full fine-tuning and the following PEFT methods:
\begin{itemize}
\item \textbf{Adapter:} \citet{houlsby2019parameter} pioneered adapter tuning (referred to in this study as \textbf{HAdapter}), which shares similar architectures to ours but injects two distinct adapters per Transformer layer. Additionally, we compare our method to \citet{pfeiffer2021adapterfusion}, who introduced a more efficient design with adapters injected only once after the FFN modules, in addition to Adapter Fusion which is used for knowledge composition among adapters in multi-task learning. However, in this study, \textbf{PAdapter} strictly refers to utilizing a single adapter per layer.
\item \textbf{LoRA~\citep{hu2021lora}:} This method freezes the pre-trained weights and injects trainable low-rank decomposition matrices ($A$ and $B$) into the attention layers to approximate weight increments.
\item \textbf{BitFit~\citep{zaken2022bitfit}:} This method modifies only the model's internal bias vectors, modifying less than 0.1\% of the total parameters.
\item \textbf{AdaLoRA~\citep{zhang2023adalora}:} An extension of LoRA that reparameterizes incremental weight updates using a singular value decomposition (SVD) formulation. AdaLoRA adaptively allocates the parameter budget across layers based on importance scoring.
\item \textbf{proPETL Adapter~\citep{zeng2023one}:} This approach shares a single prototype network across layers and tasks while jointly learning layer-specific binary masks to select distinct sub-networks. In the following experiment, the prototype network is considered as an adapter module.
\end{itemize}

\subsection{Natural Language Understanding}
\label{subsec:glue}

To evaluate SAPE's performance on encoder-only models and NLU tasks, DeBERTa-base and RoBERTa-large are fine-tuned. For the DeBERTa-base model, we choose $G=2$, resulting in two groups of adapters that each contain five layers and with both $D=16$ and $D=64$. Additionally, for RoBERTa-large, we set $G=2$ and $D=64$, resulting in two groups of 11 shared adapters across its 24 layers.

\begin{sidewaystable*}
\caption{Performance comparison of SAPE against full fine-tuning and other PEFT methods on the GLUE benchmark. Best results among PEFT methods are highlighted in bold. Baseline results for DeBERTa-v3-base are sourced from the AdaLoRA paper~\cite{zhang2023adalora}, while RoBERTa-large baseline results are reported from Xu et al.~\cite{xu2023parameter}. Results for our proposed SAPE method are independently computed.}\label{tab:glue_results}%
\begin{tabular*}{\textheight}{@{\extracolsep\fill}lcccccccccc}
\toprule
\textbf{Method} & \textbf{Params} & \textbf{MNLI} & \textbf{SST-2} & \textbf{CoLA} & \textbf{QQP} & \textbf{QNLI} & \textbf{RTE} & \textbf{MRPC} & \textbf{STS-B} & \textbf{Avg.} \\
& & (m/mm) & (Acc) & (Mcc) & (Acc/F1) & (Acc) & (Acc) & (Acc) & (Corr) & \\
\midrule
\multicolumn{11}{c}{\textit{DeBERTa-v3-base}} \\
\midrule
Full FT & 184M & 90.01 & 95.63 & 69.19 & 92.40/89.80 & 94.03 & 83.75 & 89.46 & 91.60 & 88.09 \\
\midrule
HAdapter & 0.31M & 90.06 & 95.41 & 67.65 & 91.54/88.81 & 93.52 & 83.39 & 89.25 & 91.31 & 87.60 \\
PAdapter & 0.30M & 89.97 & 94.72 & 69.06 & 91.40/88.62 & 93.87 & 84.48 & 89.71 & 91.38 & 87.90 \\
LoRA ($r=2$) & 0.33M & 90.34 & 94.95 & 68.71 & 91.61/88.91 & 94.03 & 85.56 & 89.71 & \textbf{91.68} & 88.15 \\
AdaLoRA & 0.32M & \textbf{90.68} & \textbf{95.80} & \textbf{70.04} & \textbf{91.78}/\textbf{89.16} & \textbf{94.49} & \textbf{87.36} & 90.44 & 91.63 & \textbf{88.86} \\
SAPE ($D=64, G=2$) & 0.40M & 90.10 & 95.64 & 68.89 & 91.18/88.26 & 94.03 & 86.64 & \textbf{91.18} & 91.65 & 88.48 \\
\midrule
BitFit & 0.10M & 89.64 & 94.84 & 66.96 & 88.41/84.95 & 92.24 & 78.70 & 87.75 & \textbf{91.35} & 86.02 \\
SAPE ($D=16, G=2$) & 0.10M & \textbf{89.93} & \textbf{96.10} & \textbf{68.37} & \textbf{90.33}/\textbf{87.05} & \textbf{93.79} & \textbf{87.36} & \textbf{90.20} & 91.28 & \textbf{88.22} \\
\midrule
\multicolumn{11}{c}{\textit{RoBERTa-large}} \\
\midrule
Full FT & 355M & 89.42 & 95.54 & 65.78 & 89.30/86.68 & 93.61 & 81.23 & 89.22 & 91.75 & 86.82 \\
\midrule
HAdapter & 19.7M & 91.00 & 96.37 & 67.03 & \textbf{92.19}/\textbf{88.50} & 94.31 & 85.25 & 89.94 & \textbf{92.59} & 88.35 \\
proPETL Adapter & 5.40M & \textbf{91.37} & 96.27 & 65.55 & 90.67/87.74 & \textbf{95.20} & \textbf{88.89} & 89.71 & 91.80 & 88.50 \\
AdaLoRA & 2.23M & 90.36 & 94.95 & 65.85 & 89.60/86.30 & 94.62 & 77.98 & 89.46 & 91.92 & 86.64 \\
LoRA & 1.84M & 90.76 & \textbf{96.67} & 64.47 & 90.15/86.91 & 95.00 & 79.78 & 87.50 & 91.55 & 86.78 \\
BitFit & 1.32M & 89.98 & 96.10 & 68.01 & 89.48/86.43 & 94.47 & 87.73 & 90.93 & 91.85 & 88.38 \\
SAPE ($D=64, G=2$) & 0.55M & 90.28 & 96.44 & \textbf{69.25} & 90.22/87.20 & 94.65 & 86.28 & \textbf{92.25} & 92.11 & \textbf{88.75} \\
\botrule
\end{tabular*}
\end{sidewaystable*}

\textbf{DeBERTa-base:} SAPE is evaluated against existing baselines across two distinct trainable parameter budgets, as detailed in Table~\ref{tab:glue_results}. At the $D=64$ budget (0.40M parameters), SAPE achieves an average score of 88.48\%, outperforming full fine-tuning by +0.39\%. Furthermore, it surpasses HAdapter, PAdapter, and LoRA. Although SAPE remains second to AdaLoRA in final peak accuracy at this budget,

\begin{figure}[htbp]
\centering
\includegraphics[width=\textwidth]{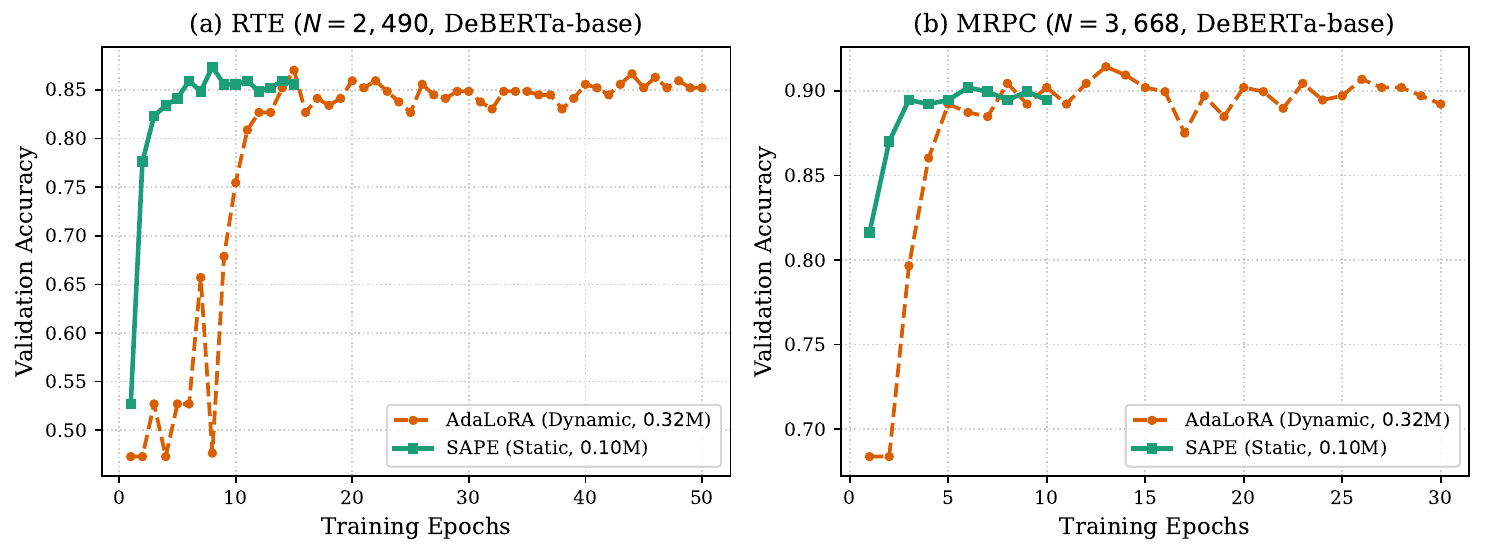} 
\caption{Epoch-wise validation trajectories for SAPE ($D=16$) and AdaLoRA across low-resource tasks. To ensure a strict and fair evaluation, AdaLoRA is implemented using the exact hyperparameter configurations and initialization schedules specified in its original paper.}\label{fig:convergence_dynamics}
\end{figure}

Figure~\ref{fig:convergence_dynamics} demonstrates that AdaLoRA's convergence speed is visibly constrained by its warmup and rank reallocation schedules. Conversely, SAPE's fixed-topology design eliminates this iterative ranking overhead entirely, allowing it to achieve significantly higher validation results during the early epochs of training.

Under the highly constrained 0.10M budget ($D=16$), SAPE outperforms BitFit by an average of +2.20\% and exceeds HAdapter, PAdapter, and LoRA while utilizing roughly one-third ($0.3\times$) the parameters. On SST-2, SAPE ($D=16$) achieves a peak accuracy of 96.10\%, outperforming all evaluated baselines alongside highly competitive performance on RTE.

\textbf{RoBERTa-large:} As reported in Table~\ref{tab:glue_results}, SAPE operates under the smallest parameter budget (0.55M) among all evaluated baselines, yet achieves the highest overall average of 88.75\%. Notably, SAPE outperforms proPETL Adapter by +0.25\%, despite proPETL requiring over $10\times$ more trainable parameters and incurring extra computational overhead for mask generation. On individual tasks, SAPE establishes new state-of-the-art results on MRPC (92.25\%) and CoLA (69.25\%), surpassing the nearest literature baselines by +1.32\% and +1.24\%, respectively.

\subsection{Natural Language Generation}
\label{subsec:llama}

We evaluate the generative and reasoning capabilities of SAPE by fine-tuning LLaMA-3.2 (3B) across three distinct tasks. For open-domain dialogue, we utilized a subset of the ConvAI2~\cite{dinan2019second} dataset (3,495 training and 100 validation dialogues), repeating each evaluation three times with different random seeds to ensure statistical robustness. Additionally, GSM8K~\cite{cobbe2021training} and CommonsenseQA~\cite{talmor2019commonsenseqa} are evaluated entirely on their standard validation partitions. SAPE is implemented with parameter budgets ranging from $\sim$0.6M to $\sim$2.5M by modulating bottleneck dimensions ($dim \in \{16, 32, 64\}$) and the number of shared groups ($G \in \{3, 4\}$). Within similar budgets, we also evaluated PAdapter, LoRA, and AdaLoRA. All methods across all benchmarks are trained for 3 epochs using FP16 precision, gradient checkpointing, and frozen backbone parameters, ensuring near-optimal performance while avoiding overfitting. Performance on ConvAI2 is measured via BLEU~\cite{papineni2002bleu}, ROUGE-L~\cite{lin2004rouge}, and BERTScore~\cite{zhang2019bertscore}, while the other two benchmarks are evaluated using Exact Match (EM) accuracy. (Note: Extended hyperparameters and evaluation details are provided in Appendix~\ref{subsec:app_llama}).

\begin{sidewaystable*}
\caption{Generative and reasoning evaluation of LLaMA-3.2 (3B) across iso-parameter budget tiers. SAPE demonstrates dominance in semantic tasks (ConvAI2, CSQA) while indicating an architectural trade-off in rigid reasoning (GSM8K).}\label{tab:llama_results}
\begin{tabular*}{\textheight}{@{\extracolsep\fill}llcccccc}
\toprule
\multirow{2}{*}{\textbf{Method}} & \multirow{2}{*}{\textbf{Topology \& Bottleneck}} & \multirow{2}{*}{\textbf{Params}} & \multicolumn{3}{c}{\textbf{ConvAI2}} & \textbf{CSQA} & \textbf{GSM8K} \\
\cmidrule(lr){4-6} \cmidrule(lr){7-7} \cmidrule(lr){8-8}
& & & BLEU & ROUGE-L & BERTScore & EM (\%) & EM (\%) \\
\midrule
Zero-Shot & Baseline & 0 & 0.0100 & 0.0947 & 0.6818 & 53.48 & 9.55 \\
\midrule
\multicolumn{8}{c}{\textit{High Budget Tier ($\sim 2.0\text{M} - 2.5\text{M}$)}} \\
\midrule
LoRA & $R=8$ & $\sim 2.29\text{M}$ & 0.0263 & \textbf{0.1563} & 0.7075 & \textbf{79.28} & 40.86 \\
PAdapter & $d=13$ & $\sim 2.32\text{M}$ & \textbf{0.0316} & 0.1636 & 0.7125 & 78.21 & \textbf{41.55} \\
AdaLoRA & final $R=7$ & $\sim 2.00\text{M}$ & 0.0239 & 0.1409 & \textbf{0.7340} & 74.20 & 32.37 \\
\textbf{SAPE (Ours)} & $G=3$, $d=64$ & $\sim 1.98\text{M}$ & 0.0275 & 0.1551 & 0.7086 & 78.87 & 40.30 \\
\textbf{SAPE (Ours)} & $G=4$, $d=64$ & $\sim 2.38\text{M}$ & 0.0262 & 0.1424 & 0.7072 & 78.54 & 38.74 \\
\midrule
\multicolumn{8}{c}{\textit{Medium Budget Tier ($\sim 1.0\text{M} - 1.5\text{M}$)}} \\
\midrule
PAdapter & $d=8$ & $\sim 1.46\text{M}$ & \textbf{0.0320} & 0.1459 & 0.7053 & 79.12 & \textbf{40.86} \\
LoRA & $R=4$ & $\sim 1.14\text{M}$ & 0.0243 & \textbf{0.1643} & 0.7150 & 78.46 & 40.71 \\
AdaLoRA & final $R=4$ & $\sim 1.14\text{M}$ & 0.0163 & 0.1320 & 0.7340 & \textbf{79.94} & 32.07 \\
\textbf{SAPE (Ours)} & $G=3$, $d=32$ & $\sim 1.00\text{M}$ & 0.0160 & 0.1467 & \textbf{0.7393} & 79.12 & 38.59 \\
\textbf{SAPE (Ours)} & $G=4$, $d=32$ & $\sim 1.20\text{M}$ & 0.0305 & 0.1528 & 0.7096 & 78.71 & 39.27 \\
\midrule
\multicolumn{8}{c}{\textit{Low Budget Tier ($\sim 0.6\text{M}$)}} \\
\midrule
LoRA & $R=2$ & $\sim 0.57\text{M}$ & 0.0186 & 0.1385 & 0.7046 & \textbf{79.12} & \textbf{39.50} \\
AdaLoRA & final $R=2$ & $\sim 0.60\text{M}$ & \textbf{0.0199} & 0.1332 & 0.7322 & 75.43 & 32.45 \\
\textbf{SAPE (Ours)} & $G=3$, $d=16$ & $\sim 0.51\text{M}$ & 0.0197 & 0.1477 & \textbf{0.7339} & 78.46 & 36.32 \\
\textbf{SAPE (Ours)} & $G=4$, $d=16$ & $\sim 0.61\text{M}$ & 0.0196 & \textbf{0.1496} & 0.7376 & 78.54 & 37.30 \\
\botrule
\end{tabular*}
\end{sidewaystable*}
($G=2$)

\textbf{ConvAI2:} The results on the ConvAI2 dataset demonstrate that SAPE surpasses prior PEFT methods in generative semantic tasks, especially in low-parameter regimes. In the highly constrained $\sim$0.6M parameter budget, SAPE (4 Groups, $dim=16$) significantly outperforms LoRA and AdaLoRA, yielding a superior ROUGE-L of 0.1496 and a BERTScore F1 of 0.7376. Furthermore, as the budget scales to the $\sim$1.0M range, SAPE (3 Groups, $dim=32$) achieves a BERTScore of 0.7393, outperforming all comparable baselines. Notably, in the high-budget tier with $dim=64$, SAPE starts to overfit due to the large bottleneck dimension. Conversely, PAdapter shows lower performance in the medium-budget tier than in the high-budget tier because its dimension must be reduced significantly, whereas SAPE maintains a suitable dimension and reduces overall parameters through cross-layer sharing. Overall, these results indicate that cross-layer parameter sharing effectively learns generalized semantic representations for conversational tasks under strict parameter constraints.

\textbf{CommonsenseQA:} This architectural efficiency strongly extends to the CommonsenseQA results. A notable observation is SAPE's ability to prevent representational degradation at extreme low-parameter constraints. While AdaLoRA suffers a catastrophic performance drop to 75.43\% at the $\sim$0.6M tier, SAPE retains a robust 78.54\% Exact Match accuracy. Moreover, the medium budget is once again a sweet spot ($\sim$1.0M) for SAPE, as it matches the 79.12\% accuracy that LoRA reaches with a double parameter budget. Overall, in world knowledge question answering, AdaLoRA achieves a better accuracy of 79.94\%, but SAPE closely chases the best result across all budgets by showing robust performance.

\textbf{GSM8K:} Conversely, evaluating SAPE on the GSM8K dataset reveals a fundamental architectural trade-off regarding rigid mathematical reasoning. While SAPE demonstrates competitive performance across all budgets and outperforms AdaLoRA in this benchmark, it is ultimately surpassed by LoRA and PAdapter. For instance, PAdapter achieves 41.55\%, whereas SAPE plateaus at 40.30\%. This reveals that while forcing layers to share bottleneck representations acts as a powerful regularizer that boosts semantic generalization, it reduces step-by-step arithmetic capability compared to non-sharing adapters. In other words, although the sandwich sharing topology shows competitive performance in this task, reaching peak accuracy in step-by-step reasoning still requires layer-wise adaptation.

\begin{figure*}[h]
\centering
\includegraphics[width=\textwidth, trim={0 0 0 25pt}, clip]{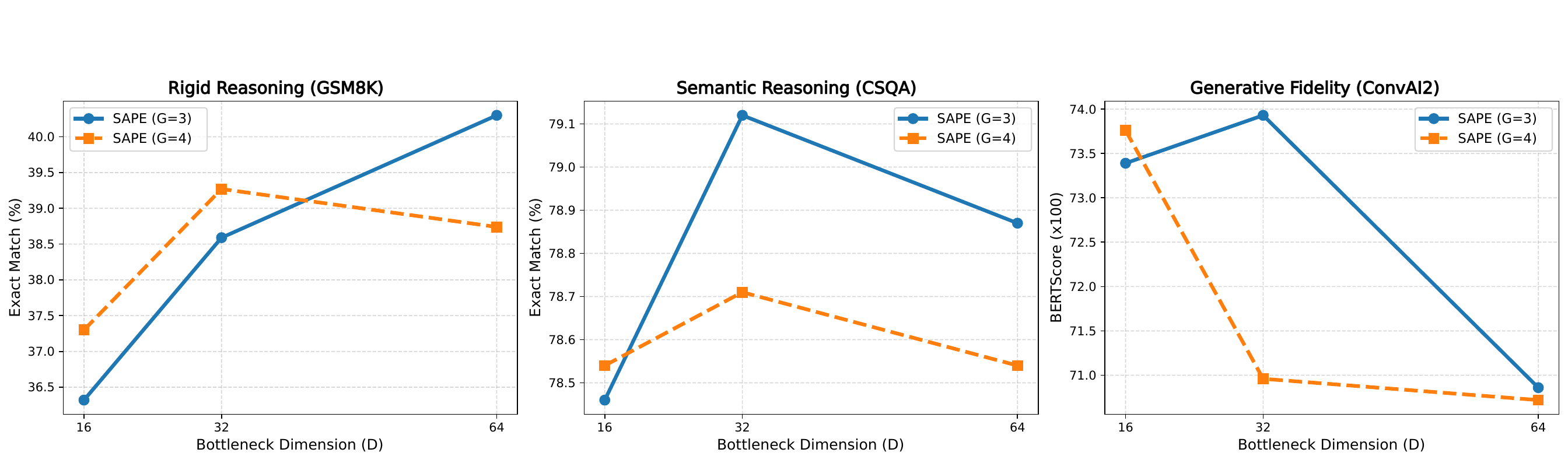}
\caption{Architectural ablation of the SAPE model comparing bottleneck capacity ($D$) and topological granularity ($G$) across rigid reasoning (GSM8K), semantic reasoning (CSQA), and generative fidelity (ConvAI2).}\label{fig:ablation_trends}
\end{figure*}

\textbf{Sensitivity Analysis of $G$ and $D$:}  As illustrated by Figure 4, scaling both variables simultaneously would lead to a higher capacity, which could lead to overfitting depending on the type of task. For instance, in rigid arithmetic reasoning (GSM8K), increasing the bottleneck dimension to $D=64$ provides a higher score, but only under a moderate group division ($G=3$). Conversely, in conversational tasks (ConvAI2) and generalized semantics (CSQA), SAPE reaches its peak performance at tighter bottlenecks ($D=16$ or $D=32$). Ultimately, a balanced configuration of $G=3$ and $D=32$ serves as the optimal architectural sweet spot. Overall, at a lower $D=16$, a higher $G=4$ yields a better result, but at a higher $D=64$, a lower $G=3$ performs better.

\subsection{Ablation Study and Empirical Analysis}
\label{subsec:ablation}

To rigorously evaluate the individual design choices within the SAPE framework, we conduct a comprehensive series of ablation studies on DeBERTa-v3-base across four representative GLUE benchmarks (RTE, QNLI, CoLA, and SST-2). We examine three core architectural dimensions: structural sharing topology, group granularity and spatial density allocation, and layer-wise dropout masking dynamics.

To ensure controlled experimental conditions, the sharing topology ablation in Section~\ref{subsec:impact_topology} evaluates all structural variants under fixed optimization parameters ($LR = 1\times10^{-4}$ with a linear decay schedule) as detailed in Appendix~\ref{subsec:app_ablation}, strictly isolating structural efficacy from hyperparameter search bias. The remaining ablations are evaluated under the framework's task-optimized hyperparameter configurations outlined in Appendix~\ref{subsec:app_deberta}.

\subsubsection{Impact of Structural Topology}
\label{subsec:impact_topology}

To understand the structural necessity of the sandwich topology, We first isolate the necessity of the sandwich topology by comparing four placement configurations, (1) \textit{Normal Sharing}: Uniformly sharing across all layers; (2) \textit{Bottom-Slice}: Isolating only the first layer ($L_0$); (3) \textit{Top-Slice}: Isolating only the final layer ($L_{N-1}$); and (4) \textit{Sandwich-Style (Ours)}: Isolating both the first and final layers.To ensure a fair comparison, four adapters with an identical bottleneck dimension ($dim=16$) were utilized across all four configurations.

\begin{table}[h]
\caption{Ablation study on adapter placement topologies across GLUE benchmark tasks using DeBERTa-base. The Sandwich-Style configuration prevents boundary interference and yields the highest average.}\label{tab:ablation}%
\begin{tabular*}{\textwidth}{@{\extracolsep\fill}lccccccc}
\toprule
\textbf{Topology} & \textbf{Distinct $L_0$} & \textbf{Distinct $L_{N-1}$} & \textbf{SST-2} & \textbf{CoLA} & \textbf{QNLI} & \textbf{RTE} & \textbf{Avg.} \\
\midrule
Normal Sharing & No & No & \textbf{95.18} & \textbf{67.65} & 49.46 & 82.31 & 73.65 \\
Bottom-Slice & Yes & No & 94.50 & 66.85 & \textbf{92.99} & 82.31 & 84.16 \\
Top-Slice & No & Yes & \textbf{95.18} & 66.80 & 81.93 & 82.67 & 81.65 \\
\textbf{Sandwich-Style} & \textbf{Yes} & \textbf{Yes} & \textbf{95.18} & 66.44 & 92.82 & \textbf{84.48} & \textbf{84.73} \\
\botrule
\end{tabular*}
\end{table}

As demonstrated in Table~\ref{tab:ablation}, the \textit{Normal Sharing} topology suffers from severe representational degradation, failing on QNLI task (49.46\%). Forcing the network to process raw lexical embeddings and deeper logical entailment using same adapter 
. Isolating the input layer (\textit{Bottom-Slice}) successfully restores QNLI performance to 92.99\%. Ultimately, the \textit{Sandwich-Style} topology establishes the optimal architectural equilibrium. By concurrently isolating the input embeddings and the task-specific output projections, it successfully prevents boundary interference while routing general semantic representations through the shared intermediate adapters.

\subsubsection{Impact of Group Granularity ($G$) and Density Allocation}

We analyze performance sensitivity with respect to the number of intermediate shared groups $G \in \{1, 2, 4, 5, 7, 10\}$, which directly governs the total trainable parameter complexity $P_{\text{total}} = (2 + G) \times (2md + m + d)$. In addition, for setups where residual layers exist such that $r = (N - 2) \bmod G \neq 0$ (specifically when $G=4$ and $G=7$), the allocation logic assigns non-uniform capacity toward either the initial input layers ($\text{DenseEarly}$) or deeper output layers ($\text{DenseLate}$). We empirically evaluate this structural density distribution as well. Throughout this experiment, the inner bottleneck dimension is held constant at $d=16$.

\begin{table}[h]
\caption{Ablation on intermediate group granularity ($G$) and structural density allocation across GLUE benchmarks.}\label{tab:ablation_groups}
\begin{tabular*}{\textwidth}{@{\extracolsep\fill}lcccccc}
\toprule
\textbf{Configuration / Group ($G$)} & \textbf{Params} & \textbf{SST-2} & \textbf{CoLA} & \textbf{QNLI} & \textbf{RTE} & \textbf{Avg.} \\
\midrule
$G=1$                      & 76K  & 95.41 & 68.81 & 93.63 & 85.92 & 85.94 \\
$G=2$                      & 101K & 96.10 & 68.37 & 93.79 & 87.36 & 86.41 \\
\addlinespace
$G=4$ (\text{DenseLate})  & 152K & 95.76 & 69.24 & 93.68 & 84.84 & 85.88 \\
$G=4$ (\text{DenseEarly}) & 152K & 96.10 & 69.27 & 93.52 & 85.56 & 86.11 \\
\addlinespace
$G=5$                      & 178K & \textbf{96.44} & 67.31 & 93.77 & 85.56 & 85.77 \\
\addlinespace
$G=7$ (\text{DenseLate})  & 228K & 95.99 & 68.28 & 93.43 & 85.56 & 85.82 \\
$G=7$ (\text{DenseEarly}) & 228K & 95.53 & 67.15 & 93.57 & 85.92 & 85.54 \\
\addlinespace
$G=10$ (PAdapter Baseline) & 304K & 95.53 & \textbf{70.08} & 93.67 & \textbf{88.44} & \textbf{86.93} \\
\botrule
\end{tabular*}
\end{table}

As illustrated in Table~\ref{tab:ablation_groups}, performance does not scale monotonically with group granularity $G$. Remarkably, $G=2$ achieves the optimal configuration among all SAPE variants, reaching an average accuracy of $86.41\%$. Specifically, SAPE with $G=2$ captures over $99.4\%$ of the performance of the full unshared baseline ($G=10$, $86.93\%$), despite using three times fewer parameters ($101\text{K}$ vs. $304\text{K}$) and dispensing with per-layer individual adapters. Notably, while the PAdapter baseline reported in Table~\ref{tab:glue_results} reflects standard published configurations from prior work, our re-tuned $G=10$ setup provides an even stronger, highly optimized baseline for unshared individual adapters under identical training conditions.

Furthermore, spatial density allocation ($\text{DenseEarly}$ vs. $\text{DenseLate}$) demonstrates marginal and mixed impact across granularities: $\text{DenseEarly}$ performs slightly better at $G=4$ ($86.11\%$ vs. $85.88\%$), whereas $\text{DenseLate}$ leads at $G=7$ ($85.82\%$ vs. $85.54\%$). This indicates that SAPE is structurally robust and does not critically depend on precise layer-density placement.

\begin{table}[h]
\caption{Comparison of parameter reduction strategies under an equivalent budget ($\sim 101\text{K}$ parameters).}\label{tab:budget_matched_ablation}%
\begin{tabular}{@{}lccccccc@{}}
\toprule
\textbf{Strategy / Method} & \textbf{Config} & \textbf{Params} & \textbf{SST-2} & \textbf{CoLA} & \textbf{QNLI} & \textbf{RTE} & \textbf{Avg.} \\
\midrule
Severe Bottleneck (PAdapter) & $G=10, d=5$ & 101K & 95.76 & 66.06 & 93.59 & 82.67 & 84.52 \\
\textbf{Aggressive Sharing (SAPE)} & $\mathbf{G=2, d=16}$ & \textbf{101K} & \textbf{96.10} & \textbf{68.37} & \textbf{93.79} & \textbf{87.36} & \textbf{86.41} \\
\botrule
\end{tabular}
\end{table}

Finally, to isolate the effect of parameter reduction mechanics under a strict parameter budget ($\sim 101\text{K}$ parameters), we compare an unshared baseline ($G=10$) using a choked bottleneck dimension ($d=5$) against SAPE ($G=2$) using a wider bottleneck dimension ($d=16$). As presented in Table~\ref{tab:budget_matched_ablation}, SAPE consistently outperforms the unshared baseline across the board, achieving a higher average accuracy ($86.41\%$ vs. $84.52\%$). 

Most notably, SAPE achieves a 4.69\% improvement on the RTE benchmark (87.36\% vs. 82.67\%). This indicates that excessively reducing the bottleneck dimension ($d=5$) limits adapter capacity on challenging entailment tasks. In contrast, SAPE maintains parameter efficiency at an equivalent budget by sharing higher-capacity adapters ($d=16, G=2$) across layers.

\subsubsection {Role of Independent Dropout Masking}

Because shared adapter modules within group $\mathcal{I}_g$ reuse identical weight matrices across distinct Transformer layers, a critical architectural decision is whether to broadcast a single static dropout mask across the entire shared group or apply random, independent masks for each layer invocation within a group. Table~\ref{tab:ablation_dropout} evaluates these strategies against a zero-dropout ($p=0.00$).

\begin{table}[h]
\caption{Ablation study on dropout masking mechanics within shared adapter layers.}\label{tab:ablation_dropout}
\begin{tabular}{@{}lcccccc@{}}
\toprule
\textbf{Masking Strategy} & \textbf{Dropout ($p$)} & \textbf{SST-2} & \textbf{CoLA} & \textbf{QNLI} & \textbf{RTE} & \textbf{Avg.} \\
\midrule
Shared Mask (across group) & 0.10 & 95.76 & 66.98 & 93.52 & 85.56 & 85.46 \\
No Dropout & 0.00 & 95.99 & 66.64 & 93.33 & 83.75 & 84.93 \\
\textbf{Independent Mask (SAPE)} & \textbf{0.10} & \textbf{96.10} & \textbf{68.37} & \textbf{93.79} & \textbf{87.36} & \textbf{86.41} \\
\botrule
\end{tabular}
\end{table}

As indicated in Table~\ref{tab:ablation_dropout}, the proposed \textbf{Independent Masking} strategy achieves superior performance across all four evaluated benchmarks, yielding an average accuracy of $86.41\%$ (+1.48\% over zero dropout). Applying identical dropout masks across shared layer adapters forces coupled co-adaptation between shared instances. Independent masks break this intra-group dependency, forcing the single shared weight matrix $\theta_g$ to learn representations resilient to diverse layer-wise noise configurations. Notably, static \textbf{Shared Masking} degrades performance relative to zero dropout on SST-2 ($95.76\%$ vs. $95.99\%$), demonstrating that shared mask broadcasting can lead to rigid spatial constraints rather than effective stochastic regularization.

\section{Conclusion and Future Work}
\label{sec:conclusion}
 
In this paper, we introduce SAPE, a highly efficient Parameter-Efficient Fine-Tuning method designed to optimize adapter-based learning. By utilizing a sandwich-style structure, we isolate boundary transformations at the initial and final layers from a shared group of intermediate adapters. This topology not only prevents destructive gradient interference but also acts as a powerful structural regularizer. Furthermore, SAPE is a highly versatile framework across a wide range of NLP tasks. At extremely constrained parameter budgets, SAPE delivers state-of-the-art efficiency in natural language understanding, open-domain generative tasks (ConvAI2), and semantic world knowledge retrieval (CommonsenseQA), successfully surpassing even dynamic methods by completely eliminating their computational overhead. Furthermore, while SAPE maintains competitive functionality across all evaluated benchmarks, we acknowledge a distinct architectural trade-off. Because cross-layer parameter sharing inherently generalizes features, it marginally smooths the sharp transformations required for highly rigid tasks. Consequently, in multi-step arithmetic reasoning (GSM8K), SAPE yields respectable results but is ultimately outperformed by unshared baselines.

Future work will explore the deployment of the SAPE framework within Federated Learning (FL) environments. Given the critical communication bottlenecks inherent to FL, SAPE's drastically minimized parameter footprint provides a significant structural advantage. Finally, extending the boundary-isolated sandwich topology beyond standard adapters to other PEFT methods such as developing a boundary-aware Shared LoRA architecture presents a compelling direction for future parameter-efficient research.

\section*{Statements and Declarations}

\textbf{Funding}\\
The authors did not receive support from any organization for the submitted work.

\textbf{Competing Interests}\\
The authors have no relevant financial or non-financial interests to disclose.

\textbf{Data Availability}\\
The datasets analyzed during the current study (GLUE, ConvAI2, CommonsenseQA, and GSM8K) are available in the public domain.

\bibliography{refs}

\begin{appendices}

\section{Comprehensive Hyperparameter Specifications}\label{sec:appendix_hyperparameters}

To ensure complete experimental transparency and reproducibility, the detailed hyperparameter configurations, optimization dynamics, and decoding setups utilized across all evaluated benchmarks and model backbones are provided below.

% ---------------------------------------------------------------------
\subsection{GLUE Benchmark Setup: DeBERTa V3-Base}\label{subsec:app_deberta}

Table~\ref{tab:hyperparameters} outlines the task-specific optimization parameters for our primary GLUE experiments under both the $D=16$ and $D=64$ bottleneck constraints.

\begin{sidewaystable*}
\caption{Detailed hyperparameter configurations for SAPE fine-tuning on \textit{DeBERTa V3-base} across all GLUE benchmark tasks. Across all experiments, the random seed is fixed to $42$, the learning rate schedule follows a cosine annealing curve, and early stopping is applied with a patience budget of $2$ epochs.}\label{tab:hyperparameters}
\begin{tabular*}{\textheight}{@{\extracolsep\fill}lcccccc@{}}
\toprule
\textbf{Task} & \textbf{Learning Rate} & \textbf{Weight Decay} & \textbf{Warmup Ratio} & \textbf{Max Epochs} & \textbf{Best Epoch} & \textbf{Batch Size} \\
\midrule
\multicolumn{7}{c}{\textit{Configuration: SAPE Bottleneck Dimension $D=16$}} \\
\midrule
MNLI & $5 \times 10^{-4}$ & 0.01 & 0.10 & 5 & 5 & 32 \\
QQP & $5 \times 10^{-4}$ & 0.01 & 0.10 & 5 & 5 & 32 \\
QNLI & $1.5 \times 10^{-3}$ & 0.01 & 0.06 & 5 & 3 & 32 \\
SST-2 & $1.5 \times 10^{-4}$ & 0.01 & 0.10 & 15 & 7 & 32 \\
CoLA & $5 \times 10^{-4}$ & 0.01 & 0.16 & 20 & 8 & 32 \\
STS-B & $1 \times 10^{-3}$ & 0.01 & 0.10 & 15 & 7 & 32 \\
MRPC & $8 \times 10^{-4}$ & 0.01 & 0.10 & 10 & 6 & 32 \\
RTE & $1 \times 10^{-3}$ & 0.01 & 0.10 & 15 & 8 & 32 \\
\midrule
\multicolumn{7}{c}{\textit{Configuration: SAPE Bottleneck Dimension $D=64$}} \\
\midrule
MNLI & $4 \times 10^{-4}$ & 0.01 & 0.10 & 7 & 6 & 32 \\
QQP & $5 \times 10^{-4}$ & 0.01 & 0.10 & 5 & 5 & 32 \\
QNLI & $8 \times 10^{-4}$ & 0.01 & 0.06 & 5 & 5 & 32 \\
SST-2 & $1.8 \times 10^{-4}$ & 0.015 & 0.10 & 12 & 7 & 32 \\
CoLA & $8 \times 10^{-4}$ & 0.01 & 0.10 & 10 & 8 & 32 \\
STS-B & $5 \times 10^{-4}$ & 0.01 & 0.05 & 20 & 5 & 32 \\
MRPC & $8 \times 10^{-4}$ & 0.01 & 0.10 & 10 & 6 & 32 \\
RTE & $5 \times 10^{-4}$ & 0.01 & 0.01 & 20 & 6 & 32 \\
\botrule
\end{tabular*}
\end{sidewaystable*}

% ---------------------------------------------------------------------
\subsection{GLUE Benchmark Setup: RoBERTa-Large}\label{subsec:app_roberta}

Table~\ref{tab:hyperparameters_roberta} details the experimental setup for the 355M parameter \textit{RoBERTa-large} backbone.

\begin{sidewaystable*}
\caption{Detailed hyperparameter configurations for SAPE ($D=64, G=2$) fine-tuning on the \textit{RoBERTa-large} backbone across all GLUE benchmark tasks. Across all experiments, the optimizer weight decay is held strictly constant at $0.01$, the random seed is fixed to $42$, and learning rate scheduling follows a cosine annealing curve.}\label{tab:hyperparameters_roberta}
\begin{tabular*}{\textheight}{@{\extracolsep\fill}lcccccc@{}}
\toprule
\textbf{Task} & \textbf{Learning Rate} & \textbf{Weight Decay} & \textbf{Warmup Ratio} & \textbf{Max Epochs} & \textbf{Best Epoch} & \textbf{Batch Size} \\
\midrule
MNLI & $4 \times 10^{-4}$ & 0.01 & 0.10 & 7 & 7 & 32 \\
QQP & $5 \times 10^{-4}$ & 0.01 & 0.10 & 5 & 5 & 32 \\
QNLI & $4 \times 10^{-4}$ & 0.01 & 0.06 & 5 & 5 & 32 \\
SST-2 & $3 \times 10^{-4}$ & 0.01 & 0.10 & 12 & 2 & 32 \\
CoLA & $1 \times 10^{-3}$ & 0.01 & 0.30 & 15 & 6 & 32 \\
STS-B & $8 \times 10^{-4}$ & 0.01 & 0.20 & 15 & 10 & 32 \\
MRPC & $6 \times 10^{-4}$ & 0.01 & 0.15 & 20 & 14 & 32 \\
RTE & $8 \times 10^{-4}$ & 0.01 & 0.10 & 10 & 10 & 32 \\
\botrule
\end{tabular*}
\end{sidewaystable*}

% ---------------------------------------------------------------------
\subsection{Generative \& Reasoning Benchmarks: Llama-3.2-3B}\label{subsec:app_llama}

Table~\ref{tab:llama_master_all} documents the optimization hyperparameters and inference decoding parameters applied to our causal language model experiments.

\begin{sidewaystable*}
\caption{Comprehensive hyperparameter, optimization, and decoding configurations for SAPE and baselines fine-tuning on the \textit{Llama-3.2-3B} backbone across mathematical reasoning (GSM8K), commonsense reasoning (CSQA), and persona-grounded dialogue (ConvAI2). Asterisks (*) denote implicit framework training defaults.}\label{tab:llama_master_all}
\begin{tabular*}{\textheight}{@{\extracolsep\fill}lccc@{}}
\toprule
\textbf{Hyperparameter} & \textbf{GSM8K} & \textbf{CommonsenseQA} & \textbf{ConvAI2 (PersonaChat)} \\
\midrule
\multicolumn{4}{c}{\textit{Optimization \& Training Setup}} \\
\midrule
Batch Size & 2 & 2 & 2 \\
Max Epochs & 3 & 3 & 3 \\
Learning Rate & $5 \times 10^{-5}$ * & $5 \times 10^{-5}$ * & $5 \times 10^{-5}$ * \\
Weight Decay & 0.0 * & 0.0 * & 0.0 * \\
LR Scheduler & Linear * & Linear * & Linear * \\
\midrule
\multicolumn{4}{c}{\textit{Inference \& Decoding Dynamics}} \\
\midrule
Primary Target Metric & Exact Match & Multiple Choice EM & BLEU / ROUGE / METEOR / BERTScore \\
Decoding Strategy & Greedy (\texttt{do\_sample=False}) & Greedy (\texttt{do\_sample=False}) & Nucleus Sampling (\texttt{top\_p=0.9}) \\
Sampling Temperature & -- & -- & 0.6 \\
Repetition Penalty & 1.0 & 1.0 & 1.1 \\
Max Generated Tokens & 256 & 2 & 64 \\
Evaluation Monte Carlo Seeds & $\{42\}$ & $\{42\}$ & $\{42, 123, 2026\}$ (Mean $\pm$ SD) \\
\botrule
\end{tabular*}
\end{sidewaystable*}

% ---------------------------------------------------------------------
\subsection{Topology Ablation Study Setup}\label{subsec:app_ablation}

\begin{table}[htbp]
\caption{Hyperparameter and architectural configurations for the SAPE topology ablation study on \textit{DeBERTa V3-base}. All four routing strategies (Regular, Top-Slice, Bottom-Slice, and Sandwich) were trained under identical optimization conditions to ensure a fair parameter budget. Asterisks (*) denote implicit framework defaults.}\label{tab:ablation_hyperparameters}
\begin{tabular}{@{}lc@{}}
\toprule
\textbf{Hyperparameter} & \textbf{Value} \\
\midrule
Batch Size (Train / Eval) & 32 \\
Learning Rate & $1 \times 10^{-4}$ \\
Optimizer & AdamW * \\
LR Scheduler & Linear * \\
Weight Decay & 0.0 * \\
Warmup Ratio & 0.0 * \\
Max Training Epochs & 30 \\
Early Stopping Patience & 3 Epochs \\
Evaluation Strategy & Epoch-wise \\
\botrule
\end{tabular}
\end{table}

\end{appendices}

\end{document}